\documentclass[sigconf]{acmart}
\AtBeginDocument{%
  }

\copyrightyear{2025}
\setcopyright{cc}
\setcctype{by}
\acmYear{2025}
\acmConference[SIGIR-AP 2025]{Proceedings of the 2025 Annual International
ACM SIGIR Conference on Research and Development in Information Retrieval in
the Asia Pacific Region}{December 7--10, 2025}{Xi'an, China}
\acmBooktitle{Proceedings of the 2025 Annual International ACM SIGIR
Conference on Research and Development in Information Retrieval in the Asia
Pacific Region (SIGIR-AP 2025), December 7--10, 2025, Xi'an,
China}
\acmDOI{10.1145/3767695.3769500}
\acmISBN{979-8-4007-2218-9/2025/12}

\usepackage{multirow}
\usepackage{multicol}
\usepackage{booktabs}
\usepackage{balance}
\usepackage{arydshln}
\usepackage{color}
\usepackage{makecell}
\usepackage{subcaption}

\usepackage{colortbl}
\definecolor{mygray}{gray}{.9}

\begin{document}

\title{Rowen: Adaptive Retrieval-Augmented Generation for Hallucination Mitigation in LLMs}
\author{Hanxing Ding}
\email{dinghanxing18s@ict.ac.cn}
\affiliation{
  \institution{State Key Laboratory of AI Safety, Institute of Computing Technology, Chinese Academy of Sciences}
  \city{Beijing}
  \country{China}
}

\author{Liang Pang}
\authornote{Corresponding author}
\email{pangliang@ict.ac.cn}
\affiliation{%
  \institution{State Key Laboratory of AI Safety, Institute of Computing Technology, Chinese Academy of Sciences}
  \city{Beijing}
  \country{China}
}

\author{Zihao Wei}
\affiliation{%
  \institution{State Key Laboratory of AI Safety, Institute of Computing Technology, Chinese Academy of Sciences}
  \city{Beijing}
  \country{China}
}

\author{Huawei Shen}
\affiliation{%
  \institution{State Key Laboratory of AI Safety, Institute of Computing Technology, Chinese Academy of Sciences}
  \city{Beijing}
  \country{China}
}

\author{Xueqi Cheng}
\affiliation{%
  \institution{State Key Laboratory of AI Safety, Institute of Computing Technology, Chinese Academy of Sciences}
  \city{Beijing}
  \country{China}
}
\renewcommand{\shortauthors}{Ding et al.}

\begin{abstract}
  Hallucinations present a significant challenge for large language models (LLMs). The utilization of parametric knowledge in generating factual content is constrained by the limited knowledge of LLMs, potentially resulting in internal hallucinations. While incorporating external information can help fill knowledge gaps, it also introduces the risk of irrelevant information, thereby increasing the likelihood of external hallucinations. To balance the use of parametric knowledge within LLMs and external information, in this study, we present \texttt{Rowen}, a novel framework that enhances LLMs with an adaptive retrieval augmentation process tailored to address hallucinated outputs. \texttt{Rowen} introduces a consistency-based hallucination detection module, which assesses the model's uncertainty regarding the input query by evaluating the semantic inconsistencies in various responses generated across different languages or models. When high uncertainties in the responses are detected, \texttt{Rowen} activates the retrieval of external information to rectify the model outputs. Through comprehensive empirical experiments, we demonstrate that \texttt{Rowen} surpasses the current state-of-the-art in both detecting and mitigating hallucinated content within the outputs of LLMs\footnote{Our Code: \url{https://github.com/dhx20150812/Rowen}.}.
\end{abstract}




\maketitle

\section{Introduction}
In recent years, large language models (LLMs) have demonstrated impressive abilities in natural language understanding~\cite{DBLP:conf/iclr/HendrycksBBZMSS21,DBLP:journals/corr/abs-2305-08322}, generation~\cite{DBLP:journals/corr/abs-2302-13971,alpaca}, and reasoning~\cite{DBLP:conf/iclr/0001Z0S23,wang-etal-2023-towards,DBLP:journals/corr/abs-2309-15402}. Despite their successes, it has been widely observed that even state-of-the-art LLMs often generate factually incorrect or nonsensical outputs, referred to as \textit{hallucinations}~\cite{DBLP:journals/csur/JiLFYSXIBMF23,DBLP:journals/corr/abs-2309-01219,DBLP:journals/corr/abs-2305-13534}. These unreliable outputs pose significant risks in practical deployments of LLMs.

\begin{figure}[t]
  \centering
  \includegraphics[width=\columnwidth]{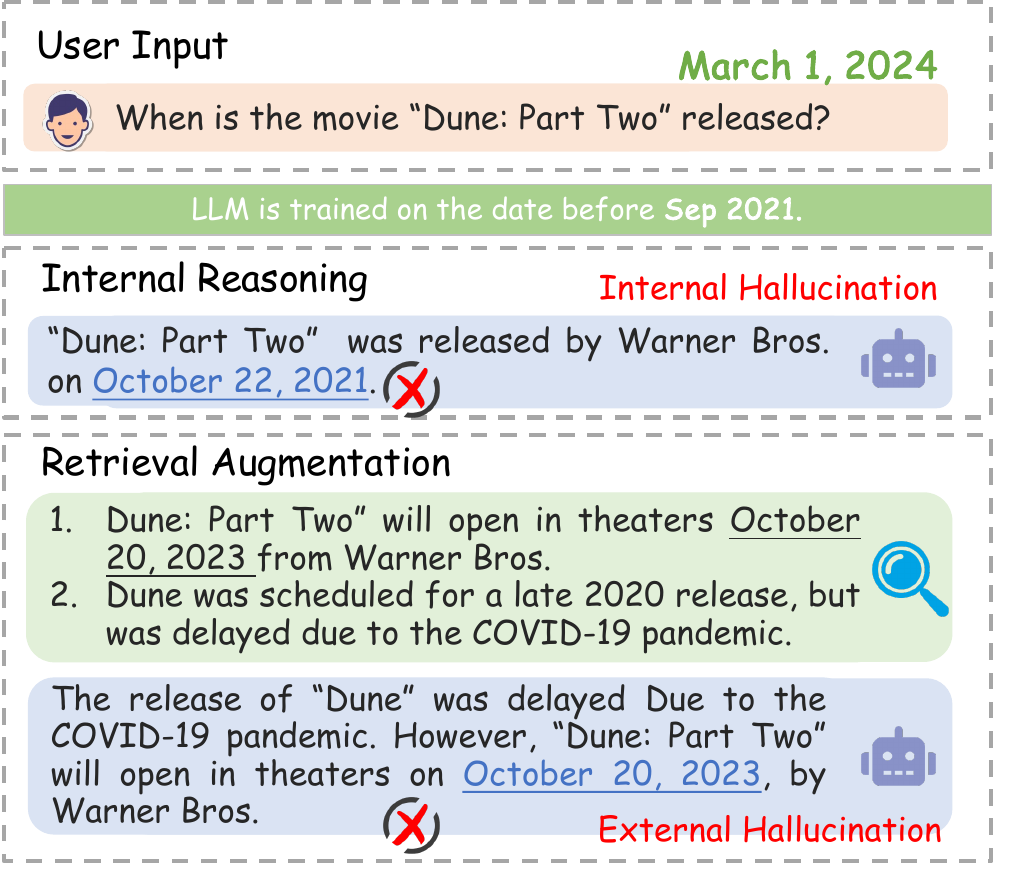}
  \caption{The limited knowledge of LLMs poses a challenge for generating accurate answers, referred to as \textit{Internal Hallucination}, when faced with the latest or domain-specific questions. Additionally, retrieval-augmented generation occasionally faces the risk of error accumulation, where irrelevant evidence may infiltrate the generation phase and lead to nonfactual responses, known as \textit{External Hallucination}.}
  \label{fig:intro}
  \vspace{-0.5cm}
\end{figure}

Efforts to enhance the factual accuracy of LLM outputs have been substantial. These studies often utilize LLMs' extensive parametric knowledge and advanced logical reasoning capabilities. They employ approaches like self-reflection~\cite{DBLP:conf/iclr/0002WSLCNCZ23,DBLP:conf/emnlp/JiYXLIF23,DBLP:journals/corr/abs-2303-17651,DBLP:journals/corr/abs-2309-11495} or collaborative refinements involving interactions among multiple models~\cite{DBLP:conf/emnlp/CohenHGG23,DBLP:journals/corr/abs-2305-14325}, aiming to enhance logical coherence in refined content. Despite their effectiveness, these self-improvement methods may be limited by LLMs' knowledge boundaries~\cite{DBLP:journals/corr/abs-2307-11019,li-etal-2023-large} or may not fully exploit parametric knowledge~\cite{DBLP:journals/corr/abs-2311-05232}, leading to what we term \textit{internal hallucination}, illustrated in Figure~\ref{fig:intro}.

Alongside these self-improvement strategies, Retrieval-Augmented Generation (RAG)~\cite{DBLP:journals/corr/abs-2307-13528} serves as a complementary method to overcome knowledge limitations. RAG employs a \textit{retrieve-then-read} pipeline~\cite{karpukhin-etal-2020-dense,DBLP:conf/nips/LewisPPPKGKLYR020}, integrating relevant documents from external knowledge sources into the LLMs' generation process~\cite{DBLP:journals/corr/abs-2311-05232,DBLP:journals/corr/abs-2309-06794,DBLP:journals/corr/abs-2307-03987,DBLP:journals/corr/abs-2307-13528,DBLP:journals/corr/abs-2305-14623}. However, as depicted in Figure~\ref{fig:intro}, RAG methods are susceptible to \textit{external hallucination} when irrelevant evidence is incorporated, potentially leading to cumulative errors and compromising output accuracy~\cite{li-etal-2023-large,DBLP:conf/icml/ShiCMSDCSZ23}.

Drawing inspiration from the latest neuroscience research~\cite{Poskanzer6538}, which reveals how the human brain dynamically switches between internal thoughts and external sensations, we introduce a novel method, termed \texttt{Rowen} (\textbf{R}etrieve \textbf{o}nly \textbf{w}h\textbf{en} it needs). \texttt{Rowen} involves an innovative consistency-based uncertainty estimation module, which perturbs semantically equivalent questions and then evaluates the semantic inconsistency of various responses across diverse languages and models when subjected to these perturbed queries. To mitigate \textit{internal hallucinations}, we trigger a retrieval process to fetch relevant information when \texttt{Rowen} detects inconsistencies in LLMs' responses, indicating internal reasoning failures. This helps LLMs refine their reasoning chains and rectify potential hallucinations. To reduce \textit{external hallucinations}, \texttt{Rowen} minimizes the risk of incorporating erroneous information by optimizing retrieval phases. If the perturbed answers convey consistent content, suggesting that LLMs are capable of generating the correct answer themselves, we directly adopt the original answer produced by internal reasoning. This method integrates parametric knowledge within LLMs with retrieved sources, ensuring a balanced integration of internal reasoning and external evidence to effectively mitigate hallucinations.

We evaluate the effectiveness of \texttt{Rowen} on the TruthfulQA dataset~\cite{lin-etal-2022-truthfulqa} and the StrategyQA dataset~\cite{geva-etal-2021-aristotle}. Remarkably, our approach excels on the TruthfulQA dataset, yielding an impressive GPT-Judge score of 59.34\%, marking a substantial improvement over the state-of-the-art (SOTA) baseline by a significant margin (+16.74\%). Similarly, on the StrategyQA dataset, our approach achieves an accuracy of 75.60\%, surpassing existing self-improvement and RAG-based baselines with notable superiority. These results underscore the powerful capability of \texttt{Rowen} in mitigating hallucinated outputs of LLMs. Furthermore, our adaptive retrieval strategy significantly reduces unnecessary retrievals, thereby enhancing the efficiency of RAG systems.

\section{Related Works}
In this section, we discuss recent works on hallucination detection and mitigation, focusing on uncertainty estimation methods for detection and post-hoc correction for mitigation.
\subsection{Exploring Uncertainty for Hallucination Detection}
Uncertainty refers to the confidence level of the model outputs, and it serves an important indicator for identifying and eliminating hallucinations, so it can assist users in determining when to trust LLMs.
In general, methods for exploring uncertainty for hallucination detection can be categorized into three types:
(1)~Logit-based estimation relies on accessing the model's logits to calculate token-level probabilities or entropy, which are used to measure uncertainty~\cite{DBLP:journals/corr/abs-2303-08896,DBLP:journals/corr/abs-2307-10236,DBLP:journals/corr/abs-2307-03987}. However, this approach can pose challenges for black-box closed-source models.
(2)~Verbalized-based estimation involves prompting language models to express their uncertainty using specific prompts~\cite{DBLP:journals/corr/abs-2306-13063,DBLP:journals/corr/abs-2305-14975,DBLP:journals/corr/abs-2305-18248}. However, these methods tend to display a high degree of overconfidence when expressing their confidence~\cite{DBLP:journals/corr/abs-2306-13063,tao2024trust}.
(3)~To overcome these limitations, consistency-based estimations are adopted to measure the consistency score among multiple responses provided by the model for a given question~\cite{DBLP:journals/corr/abs-2303-08896,DBLP:journals/corr/abs-2306-13063,DBLP:conf/iclr/0002WSLCNCZ23,zhao-etal-2023-verify}.
The underlying assumption suggests that when language models struggle with indecision and fabricate facts, they tend to provide logically inconsistent responses to identical questions.
In this work, we propose that cross-language and cross-model consistency can offer highly sensitive signals for identifying hallucinations. Therefore, we utilize cross-language and cross-model detection modules that cross-check answers to the same question across different languages or models. This cross-checking paradigm serves as a powerful mechanism to identify hallucinations in LLMs.

\subsection{Post-hoc Correction for Hallucination Mitigation}
Mitigating hallucinations in the inference time could be a cost-effective and controllable way. A line of research harnesses the extensive parametric knowledge and robust logical reasoning capabilities of LLMs to ensure logical consistency either through self-reflection within a single model~\cite{DBLP:conf/iclr/0002WSLCNCZ23,DBLP:conf/emnlp/JiYXLIF23,DBLP:journals/corr/abs-2304-14732,DBLP:journals/corr/abs-2303-17651,DBLP:journals/corr/abs-2309-11495} or through collaborative refinements or debates involving multiple models~\cite{DBLP:conf/emnlp/CohenHGG23,DBLP:journals/corr/abs-2305-14325}. Despite their strengths, LLMs are sometimes constrained by their knowledge boundaries and the complexity of the reasoning chain, resulting in occasional inaccuracies~\cite{DBLP:journals/corr/abs-2307-11019,li-etal-2023-large,mallen-etal-2023-trust} termed \textit{internal hallucination}. To address this knowledge gap, retrieval-augmented generation methods leverage external knowledge as supplementary evidence to aid LLMs in providing accurate responses~\cite{DBLP:journals/corr/abs-2311-05232,DBLP:journals/corr/abs-2309-06794,DBLP:conf/sigir/NiuGLC12,DBLP:journals/corr/abs-2402-02764,DBLP:journals/corr/abs-2307-03987,DBLP:journals/corr/abs-2307-13528,DBLP:journals/corr/abs-2305-14623}. However, these approaches, while effective, occasionally encounter the challenge of error accumulation, where irrelevant evidence may seep into the generation process, leading to incorrect responses~\cite{li-etal-2023-large,DBLP:conf/icml/ShiCMSDCSZ23}, a phenomenon referred to as \textit{external hallucination}. Our work only performs retrieval augmentation when hallucinations are detected, thereby maximizing the utilization of both the parametric knowledge and externally retrieved information.

There are also some adaptive retrieval methods that assess the difficulty of questions or the confidence in responses to decide whether to retrieve documents~\cite{DBLP:conf/emnlp/JiangXGSLDYCN23,mallen-etal-2023-trust,DBLP:journals/corr/abs-2310-11511,DBLP:journals/corr/abs-2403-14403}.

\section{Methodology}

\begin{figure*}[ht]
  \centering
  \includegraphics[width=1.85\columnwidth]{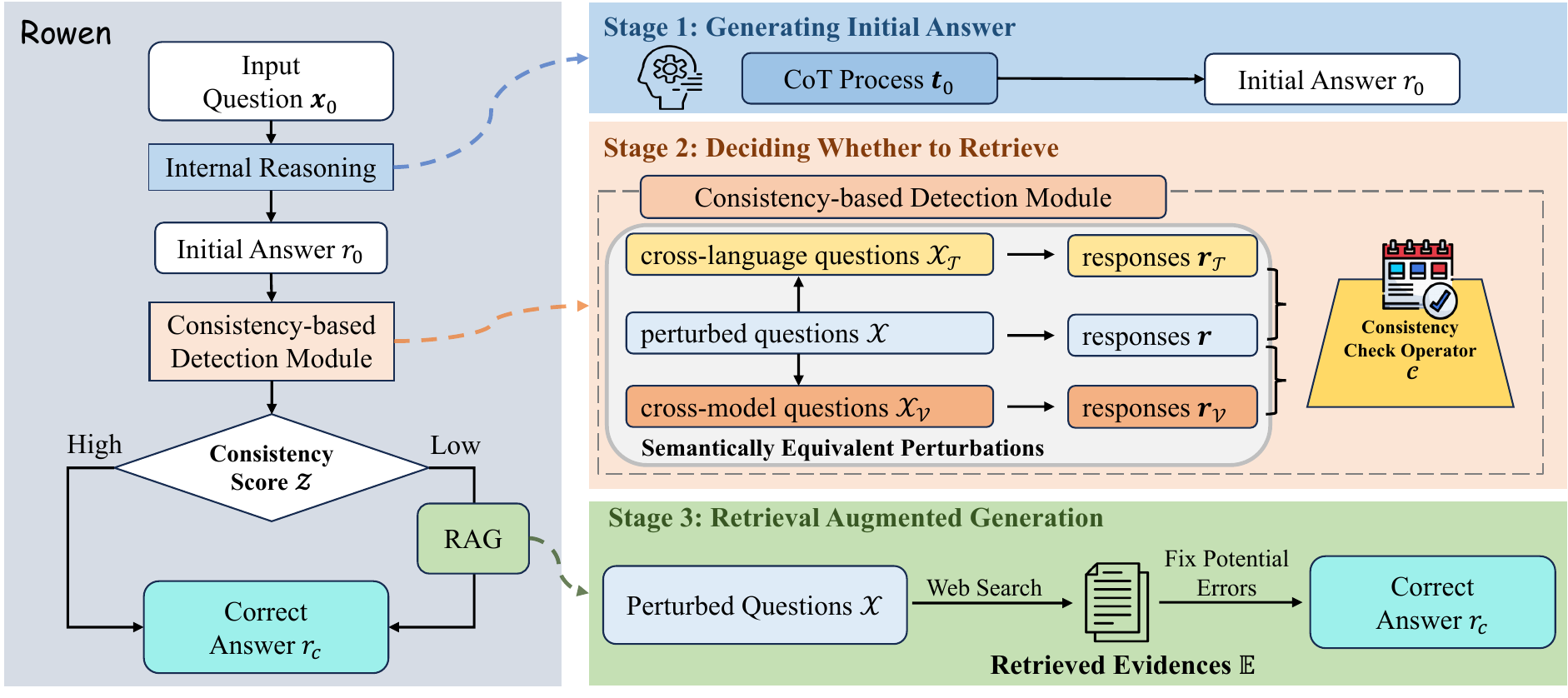}
  \caption{Overview of \texttt{Rowen} framework. We start by producing an initial response using CoT thinking. Then, a consistency-based detection module evaluates the semantic consistency of responses to the same question in different languages or models. If inconsistencies arise, a retrieval-augmented mechanism is engaged to fetch external information, helping to rectify the reasoning and correct any inaccuracies. Otherwise, the initial answer is retained.}
  \label{fig:method}
\end{figure*}

Our objective is to enhance the factuality of LLM responses by integrating parametric and external knowledge. We propose a framework called \texttt{Rowen} (\textbf{R}etrieve \textbf{o}nly \textbf{w}h\textbf{en} needed). Initially, we leverage LLMs' Chain-of-Thought (CoT) reasoning to generate an initial response (\S~\ref{sec:cot_answer}). To mitigate internal hallucinations, \texttt{Rowen} employs a consistency-based hallucination detection module that assesses the reliability of the initial response (\S~\ref{sec:multilingual_detection}). If high uncertainties are found, the initial answer is refined using external information via retrieval augmentation (\S~\ref{sec:rag}), resulting in the final response. Otherwise, the initial response is considered the final output. For external hallucinations, \texttt{Rowen} resorts to external knowledge only when high uncertainties are found, ensuring that the final answer is both accurate and reliable.

\begin{table}[ht]
\small
\centering
\begin{tabular}{p{7.2cm}}
\toprule
\texttt{When responding to my question, please first evaluate the validity of the information or the assumption underlying the question. Once you've established its truth or existence, then proceed to deliver a detailed explanation or answer. Prioritize accuracy and fact-checking before diving into elaboration or conjecture.} \\
\bottomrule
\end{tabular}
\caption{The prompt used to generate CoT answer.}
\label{tab:cot_prompt}
\vspace{-0.3cm}
\end{table}

\subsection{Stage 1: Generating Initial Answer}
\label{sec:cot_answer}
To maximize the exploitation of the parametric knowledge in LLMs, we initially employ their Chain-of-Thought (CoT) reasoning to generate a preliminary response. This process involves: critically assessing the validity of the information in the input query $\boldsymbol{x}_0$ and prioritizing accuracy and fact-checking for answer elaboration, detailed in Table~\ref{tab:cot_prompt}. After generating the CoT thought $\boldsymbol{t}_0$, we ask $\mathcal{M}$ to provide a concise answer $\boldsymbol{r}_0$ for the input query $\boldsymbol{x}_0$. Our aim is to ensure high-quality responses are generated from the outset. The answer $\boldsymbol{r}_0$ is finalized as the ultimate response to the input query $\boldsymbol{x}_0$ after our detection module ensures it is free from hallucinations.

\subsection{Stage 2: Deciding Whether to Retrieve}
\label{sec:multilingual_detection}
To decide when to retrieve, we leverage model uncertainty, which refers to the confidence level of model outputs and serves as a crucial indicator for deciding when to trust LLMs~\cite{DBLP:journals/corr/abs-2309-01219}. Unfortunately, current consistency-based methods fail when LLMs provide consistent yet incorrect answers across different perturbations~\cite{DBLP:conf/emnlp/ZhangLDMS23}.  This issue may arise because these methods focus exclusively on semantic coherence within a single language or model.

To tackle this issue, we propose novel cross-language and cross-model detection modules that assess semantic consistency among responses for the same question across different languages or models. If inconsistencies are detected in these responses, we flag them as potentially inaccurate and invoke a retrieval process.

\subsubsection{Cross-Language / Model Perturbations}
To facilitate subsequent consistency-based hallucination detection, we begin by leveraging advancements in LLM prompting to generate semantically equivalent perturbations. Initially, we start with an input $\boldsymbol{x}_0$ and instruct the model $\mathcal{M}$ to provide a set of $k$ semantically equivalent questions $\mathcal{X}=\left\{\boldsymbol{x}^{1}, \boldsymbol{x}^{2}, \ldots, \boldsymbol{x}^{k}\right\}$. We use the prompt: "\texttt{For the question [ORIGINAL QUESTION], please provide {k} semantically equivalent questions}" with a high decoding temperature to generate diverse perturbed expressions.

After obtaining the diverse verbalized questions, we prompt the LM $\mathcal{M}$ to generate its candidate responses according to the questions. We employ a greedy decoding strategy to avoid unpredictable randomness of the LM $\mathcal{M}$ as much as possible.
\begin{equation}
  \begin{aligned}
    \boldsymbol{r}^{j} = \mathcal{M}(\boldsymbol{x}^{j}), j=1,\ldots,k \; ,
  \end{aligned}
\end{equation}
where $k$ is the length of the generated semantically equivalent questions $\mathcal{X}$.

\paragraph{Cross-Language Detection}
To capture language-level uncertainty, we first incorporate a cross-language consistency check (\texttt{Rowen-CL}) to evaluate the semantic consistency of responses to the same question across different languages. To achieve this, we introduce language-level perturbations by asking the model $\mathcal{M}$ to translate the source-language questions $\mathcal{X}$ into corresponding paraphrased questions $\mathcal{X}_{\mathcal{T}}=\left\{\boldsymbol{x}_{\mathcal{T}}^{1}, \boldsymbol{x}_{\mathcal{T}}^{2}, \ldots, \boldsymbol{x}_{\mathcal{T}}^{k}\right\}$ in the target language. The model $\mathcal{M}$ is then instructed to generate corresponding answers to each question in the target language.
\begin{equation}
  \begin{aligned}
    \boldsymbol{r}_{\mathcal{T}}^{j} = \mathcal{M}(\boldsymbol{x}_{\mathcal{T}}^{j}), \; j=1,\ldots,k.
  \end{aligned}
\end{equation}

\paragraph{Cross-Model Detection}
Besides language-level cross-checking, we also introduce a cross-model detection module (\texttt{Rowen-CM}) to evaluate the semantic consistency of responses to the same question across different models. We adopt an additional verifier LM $\mathcal{M}_{\mathcal{V}}$ for model-level cross-checking and instruct the verifier LM $\mathcal{M}_{\mathcal{V}}$ to provide answers for each source question:
\begin{equation}
  \begin{aligned}
    \boldsymbol{r}_{\mathcal{V}}^{j} = \mathcal{M}_{\mathcal{V}}(\boldsymbol{x}^{j}), \; j=1,\ldots,k.
  \end{aligned}
\end{equation}

\subsubsection{Consistency Score Calculation}
In this step, we utilize the generated questions and answers from all previous stages to calculate a numerical consistency score that captures language-level and model-level cross-checking uncertainties.

\begin{table}[ht]
  \small
  \centering
  \begin{tabular}{p{7.2cm}}
  \toprule
  \texttt{Given the question Q, and two potential answers: answer A in English and answer B in Chinese. Your task is to determine if the content and meaning of A and B are equivalent in the context of answering Q. Consider linguistic nuances, cultural variations, and the overall conveyance of information. Respond with a binary classification. If A and B are equivalent, output 'True.', otherwise output 'False'} \\
  \bottomrule
  \end{tabular}
  \caption{The instruction for determining whether two QA pairs in different languages are semantically equivalent.}
  \label{tab:language_consistency_prompt}
  \vspace{-0.3cm}
\end{table}

\paragraph{Cross-Language Consistency Score} Let $\mathcal{C}(\cdot, \cdot)$ denote a semantic equivalence checking operator that takes two QA pairs as inputs and returns "\texttt{True}" if they are semantically equivalent, and "\texttt{False}" otherwise. We then map the response to a numerical semantic equivalent score: \texttt{\{"True" $\rightarrow$ 1.0, "False" $\rightarrow$ 0.0\}}. In our implementation, we leverage the model $\mathcal{M}$ and utilize the prompt provided in Table~\ref{tab:language_consistency_prompt} to implement the cross-language checking operator and calculate the cross-language consistency score $\mathcal{Z}$ as:
\begin{equation}
  \begin{aligned}
    \mathcal{Z}_{\text{CL}} = \frac{1}{k}\sum_{j=1}^{k}\mathcal{C}(\mathcal{P}^{j}, \mathcal{P}_{\mathcal{T}}^{j}),
  \end{aligned}
\end{equation}
where $\mathcal{P}=(\boldsymbol{x},\boldsymbol{r})$ and $\mathcal{P}_{\mathcal{T}}=(\boldsymbol{x}_{\mathcal{T}},\boldsymbol{r}_{\mathcal{T}})$ denote the QA pairs in the source language and target language, respectively.

\begin{table}[htbp]
  \small
  \centering
  \begin{tabular}{p{7.2cm}}
  \toprule
  \texttt{Are the following two Question-Answer(QA) pairs semantically equivalent? Provide your best guess that it is correct (True or False). Given ONLY the guess (True or False), no other words or explanation.} \\
  \bottomrule
  \end{tabular}
  \caption{The instruction for determining whether two QA pairs generated by different models are semantically equivalent.}
  \label{tab:model_consistency_prompt}
  \vspace{-0.3cm}
\end{table}

\paragraph{Cross-Model Consistency Score} Similar to the cross-language consistency score calculation, we use the prompt in Table~\ref{tab:model_consistency_prompt} to implement the checking operator $\mathcal{C}$ to calculate cross-model consistency score:
\begin{equation}
  \begin{aligned}
    \mathcal{Z}_{\text{CM}} = \frac{1}{k}\sum_{j=1}^{k}\mathcal{C}(\mathcal{P}^{j}, \mathcal{P}_{\mathcal{V}}^{j}),
  \end{aligned}
\end{equation}
where $\mathcal{P}_{\mathcal{V}}=(\boldsymbol{x}_{\mathcal{V}},\boldsymbol{r}_{\mathcal{V}})$ denote the QA pairs generated by verifier model $\mathcal{M}_{\mathcal{V}}$.

\paragraph{Hybrid Consistency Score}
The different variants of \texttt{Rowen} capture various aspects of uncertainty in the original response, complementing each other effectively. We propose integrating the cross-language and cross-model consistency scores to create a unified hybrid consistency score:
\begin{equation}
  \begin{aligned}
    \mathcal{Z}_{\text{Hybrid}} = \mathcal{Z}_{\text{CL}} + \alpha * \mathcal{Z}_{\text{CM}},
  \end{aligned}
\end{equation}
where $\alpha$ is a weight factor for the cross-model consistency score.

\subsection{Stage 3: Retrieval Augmented Generation}
\label{sec:rag}
If the consistency score $\mathcal{Z}$ falls below a threshold, it indicates possible hallucinated content in the original response $\boldsymbol{r}_0$. We then introduce a retrieval-augmented generation procedure.

\paragraph{Searching Relevant Knowledge}
To help the LM $\mathcal{M}$ correct errors, we search for supporting evidence from external sources like online webpages. We first ask the model $\mathcal{M}$ to generate search queries for each paraphrased question in $\mathcal{X}$. These queries are input into the online search engine to retrieve relevant knowledge, denoted as $\mathbf{E}$, used for correcting factual errors in $\boldsymbol{r}_0$.

\paragraph{Repairing Hallucinated Contents}
With the retrieved evidence $\mathbf{E}$, the model reviews the original thought process $\boldsymbol{t}_0$ and initial answer $\boldsymbol{r}_0$. The aim is to identify and correct inaccuracies, producing the refined answer $\boldsymbol{r}_{c}$:
\begin{equation}
\begin{aligned}
\boldsymbol{r}_{c} = \mathcal{M}(\boldsymbol{x}_0, \boldsymbol{t}_0, \boldsymbol{r}_0, \mathbf{E}).
\end{aligned}
\end{equation}
The corrected answer $\boldsymbol{r}_{c}$ serves as the final response to question $\boldsymbol{x}_0$.

\begin{table*}[t]
    \centering
    \resizebox{0.7\linewidth}{!}{
    \begin{tabular}{lrrrr}
    \toprule
    \multicolumn{1}{l}{\multirow{2}{*}{\textbf{Models}}} & \multicolumn{3}{c}{\textbf{TruthfulQA}} & \textbf{StrategyQA} \\
    \cmidrule(lr){2-4} \cmidrule(lr){5-5}
    \multicolumn{1}{r}{}  & \textbf{GPT-Judge $\uparrow$}   & \textbf{BLEU $\uparrow$} & \textbf{Rouge-L $\uparrow$}  & \textbf{Accuracy $\uparrow$}  \\
    \midrule
    \rowcolor{mygray}
    \textit{Vanilla LLMs} & & & & \\
    \quad ChatGPT (\texttt{gpt-3.5-turbo})   &    47.92  &  10.17    &    \textbf{31.31}     &    61.40     \\
    \midrule
    \rowcolor{mygray}
    \textit{Self-improvement Methods} & & & & \\
    \quad CoVe~\cite{DBLP:journals/corr/abs-2309-11495}     & 48.01    & 12.81   & 26.52       &   61.40    \\
    \quad Multi-agent Debate~\cite{DBLP:journals/corr/abs-2305-14325}   & 50.83       & 3.94      & 21.05    &  65.73   \\
    \quad Self-Reflection~\cite{DBLP:conf/emnlp/JiYXLIF23}  &  42.99  &  3.86  &  18.18  & 62.40 \\
    \midrule
    \rowcolor{mygray}
    \textit{Retrieval-augmented Methods} & & & & \\
    \quad Factool~\cite{DBLP:journals/corr/abs-2307-13528}    & 34.50      & 1.34    & 12.22   &  67.20 \\
    \quad Detect-and-Mitigate~\cite{DBLP:journals/corr/abs-2307-03987}   & 49.98    & 3.17     & 18.59   &  56.94 \\
    \midrule
    \rowcolor{mygray}
    \textit{Adaptive Retrieval Methods} & & & & \\
    \quad FLARE~\cite{DBLP:conf/emnlp/JiangXGSLDYCN23}   & 45.04   & 11.59     & 26.83   &  61.19 \\
    \quad Adaptive-Retrieval~\cite{mallen-etal-2023-trust}   & 45.55   & 8.87     & 26.75   &  62.50 \\
    \quad Self-RAG~\cite{DBLP:journals/corr/abs-2310-11511}   & 40.36   & 4.36     &  21.28   &  58.40 \\
    \quad Adaptive-RAG~\cite{DBLP:journals/corr/abs-2403-14403}   & 46.02   & 10.29     &  26.24   &  68.50 \\
    \quad LUQ~\cite{DBLP:journals/corr/abs-2403-20279} & 55.08 & 5.79 & 21.44 & 71.00 \\
    \midrule
    \rowcolor{mygray}
    \textit{Our Framework} & & & & \\
    \quad \texttt{Rowen-CL}   & 57.39 & 7.60 & 24.16 &  74.00 \\
    \quad \texttt{Rowen-CM} & 56.29 & 6.85 & 22.36 &  72.40 \\
    \quad \texttt{Rowen-Hybrid} & \textbf{59.34} & \textbf{15.27} & 31.15 &  \textbf{75.60} \\
    \bottomrule
    \end{tabular}}
    \caption{Experimental results of mitigating hallucinations on the TruthfulQA dataset and StrategyQA dataset. \texttt{Rowen-Hybrid} achieves a detection accuracy of 59.0\% on the TruthfulQA dataset and 73.0\% on the StrategyQA dataset.}
    \label{tab:main_result}
    \vspace{-0.5cm}
\end{table*}

\section{Experimental Setup}
In this section, we outline the experimental setup of \texttt{Rowen}, detailing the datasets utilized, the evaluation metrics employed, the baseline methods considered, and the implementation specifics.

\subsection{Datasets and Evaluation Metrics}
We evaluate the hallucination mitigation performance on the following datasets:
\paragraph{TruthfulQA}
We use the TruthfulQA dataset~\cite{lin-etal-2022-truthfulqa} to evaluate the ability of LLMs in generating truthful responses~\cite{DBLP:journals/corr/abs-2312-15710,DBLP:journals/corr/abs-2207-05221}. In our study, we focus on the generation task in TruthfulQA. Therefore, to evaluate the factuality of responses from LLMs, we calculate the GPT-Judge score, obtained by fine-tuning the \texttt{babbage-002} model using the original fine-tuning data from their official repository\footnote{\url{https://github.com/sylinrl/TruthfulQA}}. We also report the BLEU and Rouge-L scores to evaluate the lexical overlap between generated responses and ground-truth references.
\paragraph{StrategyQA}
The StrategyQA dataset~\cite{geva-etal-2021-aristotle} comprises crowdsourced \texttt{yes} / \texttt{no} questions that require multi-step reasoning for accurate answers. We follow previous work~\cite{DBLP:conf/emnlp/JiangXGSLDYCN23} to randomly sample 500 examples due to the cost consideration of running experiments. We also follow the settings of \citet{DBLP:conf/nips/Wei0SBIXCLZ22} to generate both the reasoning process as well as the final answer. We present the exact-match accuracy of the generated \texttt{yes} / \texttt{no} answers compared to the gold-standard answers.

\subsection{Baseline Methods}
We consider the following methods as our baselines: (1) Vanilla LLMs, such as ChatGPT.
(2) Self-improvement methods: \textbf{CoVe}~\cite{DBLP:journals/corr/abs-2309-11495} generates verification questions to self-analyze potential errors, systematically addressing each question to refine the baseline response. \textbf{Self-Reflection}~\cite{DBLP:conf/emnlp/JiYXLIF23} presents an interactive self-reflection methodology that incorporates knowledge acquisition and answer generation.
\textbf{Multi-agent Debate}~\cite{DBLP:journals/corr/abs-2305-14325} utilize multiple LM agents to debate their individual responses over multiple rounds to arrive at a common final answer.
(3) Retrieval-augmented methods: \textbf{Factool}~\cite{DBLP:journals/corr/abs-2307-13528} leverages various tools to gather evidence about the factuality of the generated content. \textbf{Detect and Mitigate}~\cite{DBLP:journals/corr/abs-2307-03987} actively detects hallucinations during generation by identifying potential hallucination through the logit output values of LLMs.
(4) Adaptive retrieval methods: \textbf{FLARE}~\cite{DBLP:conf/emnlp/JiangXGSLDYCN23} adopts an active retrieval strategy that only retrieves when LLMs generate low probability tokens. \textbf{Adaptive-Retrieval}~\cite{mallen-etal-2023-trust} only retrieves when necessary based on a pre-defined threshold for entity popularity. \textbf{Self-RAG}~\cite{DBLP:journals/corr/abs-2310-11511} trains a single arbitrary LM that adaptively retrieves passages on-demand. \textbf{Adaptive-RAG}~\cite{DBLP:journals/corr/abs-2403-14403} trains a query-complexity classifier to decide when to retrieve based on question complexity. \textbf{LUQ}~\cite{DBLP:journals/corr/abs-2403-20279} is a sampling-based uncertainty quantification for long text.

\subsection{Implementation Details}
\paragraph{LLMs} In our experiments, we validate \texttt{Rowen} using the GPT-3.5 language model, specifically the \texttt{gpt-3.5-turbo-0613} version\footnote{\url{https://api.openai.com/v1/chat/completions}}. We re-implement all baselines, except Self-RAG, using GPT-3.5 to ensure a fair comparison.
For semantic perturbations, we configure the temperature to 1.0 to generate diverse perturbed expressions.
Otherwise, the temperature is set to 0.0 to obtain high-quality deterministic outputs.
Considering the diversity of expressions and the latency in generating perturbations, we produce $k=6$ semantically equivalent questions.
For the cross-language detection module, English serves as the source language while Chinese is employed as the target language\footnote{We study the impact of the choice of target language in \S~\ref{sec:language_pair} Table~\ref{tab:language}.}.
This decision is made considering the substantial cultural disparities between the two languages, which can enhance the model's capability to detect semantic inconsistencies when responding to identical questions.
For the cross-model detection module, we adopt \texttt{Qwen-Max-0428}\footnote{\url{https://qwenlm.github.io/blog/qwen-max-0428}}, a large instruction-tuned model for chat service, as the verifier LM\footnote{We also study the impact of the choice of verifier LM in \S~\ref{sec:verifier_lm} Table~\ref{tab:model}.}.
We execute all experiments on 8 NVIDIA A800 80G GPUs.

\begin{table*}[t]
    \centering
    \resizebox{0.85\linewidth}{!}{
    \begin{tabular}{lrrrrrr}
    \toprule
    \multicolumn{1}{l}{\multirow{2}{*}{\textbf{Models}}} & \multicolumn{4}{c}{\textbf{TruthfulQA}} & \multicolumn{2}{c}{\textbf{StrategyQA}} \\
    \cline{2-5}\cline{6-7}
    \multicolumn{1}{c}{}  & GPT-Judge $\uparrow$  & BLEU $\uparrow$ & Rouge-L $\uparrow$ & Ratio(\%) & Accuracy $\uparrow$ & Ratio(\%) \\
    \midrule
    \rowcolor{mygray}
    \textit{LLaMa2-7B} & & & & & & \\
    Average Token Probability  & 50.19  &  12.51 &  31.11 &  46.5 & 71.50 & 23.0 \\
    Average Token Entropy &  52.22  & 9.18 & 28.37 &  59.0 & 72.00 & 26.5 \\
    \midrule
    \rowcolor{mygray}
    \textit{SelfCheckGPT} & & & & & & \\
    w/ BERTScore  & 51.38 & 8.43 & 26.85 & 27.2 & 67.50 & 21.0 \\
    w/ MQAG   & 52.76 & 7.69 & 26.92 &  54.4 & 69.00 & 34.0 \\
    w/ Ngram   & 52.41 & 5.01 & 21.60 &  34.9 & 66.50 & 32.0 \\
    Combination   & 53.10 & 6.69 & 24.04 &  51.3 & 69.50 & 30.0 \\
    \midrule
    \rowcolor{mygray}
    \textit{Consistency} & & & & & & \\
    \texttt{SAC$^3$-Q} & 51.02 & 7.90 & 28.00 & 24.5 & 65.50 & 24.0 \\
    \texttt{SAC$^3$-all} & 52.22 & 9.37 & 29.56 &  20.8 & 67.00 & 24.5 \\
    \midrule
    \texttt{Rowen-Hybrid}   & \textbf{59.34} & \textbf{15.27} & \textbf{31.15} &  23.0 & \textbf{75.60} & 20.0  \\
    \bottomrule
    \end{tabular}}
    \caption{Performance comparison of applying other hallucination detection methods in adaptive retrieval scenarios. We also report the ratio of retrieval conducted by each method.}
    \label{tab:detection}
    \vspace{-0.5cm}
\end{table*}

\paragraph{Retrieval Module} Following ~\citet{DBLP:journals/corr/abs-2307-13528}, we utilize the Google Search API offered by Serper\footnote{\url{https://serper.dev/}} to search the top pages and extract the most pertinent search snippets from the API's response. 
Subsequently, we parse the response to acquire various types of snippets, including answer boxes, knowledge graphs, and organic search results.

\section{Experimental Results}
In this section, we evaluate the effectiveness of \texttt{Rowen} for hallucination mitigation. In particular, we aim to answer the following research questions:
\begin{itemize}
    \item \textbf{RQ1}: How does \texttt{Rowen} perform compared to existing strong methods for hallucination mitigation?
    \item \textbf{RQ2}: Is the proposed multilingual detection module superior to other hallucination detection methods?
    \item \textbf{RQ3}: How well does \texttt{Rowen} generalize to new datasets and models?
    \item \textbf{RQ4}: How do different hyperparameters affect the performance of \texttt{Rowen}?
    \item \textbf{RQ5}: Does \texttt{Rowen} eliminate both types of hallucinations as expected?
    \item \textbf{RQ6}: Does \texttt{Rowen} reduce unnecessary retrieval calls?
\end{itemize}

\subsection{Main Results (RQ1)}
We evaluate the effectiveness of \texttt{Rowen} on the TruthfulQA and StrategyQA datasets. Table~\ref{tab:main_result} presents the overall performance of \texttt{Rowen} compared to several strong baselines. \texttt{Rowen} demonstrates superior performance on both datasets, with higher GPT-Judge score and accuracy, indicating the effectiveness of our proposed method.

Vanilla ChatGPT shows a certain level of accuracy in answering factual questions, achieving scores of 47.92\% and 61.40\% on the two datasets, respectively. While self-improvement methods perform better than the vanilla LM on both datasets, they are still limited by their knowledge boundaries and suffer from \textit{internal hallucinations}.  RAG methods demonstrate relatively better performance compared to self-improvement methods, highlighting the benefits of integrating external knowledge. However, Factool falls short on the TruthfulQA dataset, and the Detect-and-Mitigate method underperforms on the StrategyQA dataset. This may be attributed to error accumulation caused by unnecessary retrieval (\textit{external hallucinations}).

We also conduct additional experiments to compare with four adaptive retrieval methods. Notably, Adaptive-Retrieval faces challenges on the TruthfulQA dataset due to some questions lacking explicit entities, causing it to struggle in deciding when to retrieve based on entity popularity, leading to poor performance. Besides, Self-RAG's effectiveness is hindered by the capabilities of the LLaMa model, resulting in inferior performance.

Compared to the aforementioned baselines, \texttt{Rowen} demonstrates significant performance gains on both datasets. Both \texttt{Rowen-CL} and \texttt{Rowen-CM} exhibit excellent hallucination mitigation capabilities, even when compared to strong adaptive retrieval methods. Specifically, \texttt{Rowen-Hybrid} achieves a GPT-Judge score of 59.34\% on the TruthfulQA dataset, surpassing the strongest baseline by 16.74\%. Additionally, \texttt{Rowen-Hybrid} attains an accuracy of 75.60\% on the StrategyQA dataset, significantly outperforming existing baselines. These results underscore \texttt{Rowen}'s ability to effectively leverage parametric knowledge and external information for maximum advantage.

\subsection{Effect of Detection Module (RQ2)}
To validate the effectiveness of our proposed hallucination detection module, we compare its performance in adaptive retrieval scenarios for hallucination mitigation with strong detection methods: (1) average token-level probability / entropy that utilizes the probabilities / entropies of tokens generated by a proxy LM (e.g., LLaMa2-7B) as a metric to measure hallucination. (2) SelfCheckGPT~\cite{DBLP:journals/corr/abs-2303-08896} that measures information consistency between the different responses to determine hallucinations. (3) Consistency-based method~\cite{DBLP:conf/emnlp/ZhangLDMS23}, \texttt{SAC$^3$}, that evaluates semantic-aware cross-check consistency, building upon the foundation of self-consistency principles.

Based on the results in Table~\ref{tab:detection}, it is evident that logits-based methods perform moderately well in detecting hallucinations. Specifically, the entropy-based estimation exhibits superior performance. 
Besides, the combination of different variants in SelfCheckGPT leads to a slight performance improvement. Additionally, it is worth noting that \texttt{SAC$^3$} achieve competitive performance on detecting potential factual errors accurately.

Finally, we observe that \texttt{Rowen} significantly outperforms these strong hallucination detection baselines, especially the monolingual detection method \texttt{SAC$^3$}. We also notice that \texttt{Rowen} achieves notable hallucination mitigation with a minimal number of retrieval calls. This underlines the superior efficiency of our adaptive retrieval module.

\subsection{Scalability of \texttt{Rowen} (RQ3)}

\begin{figure}[t]
    \centering
    \includegraphics[width=1\columnwidth]{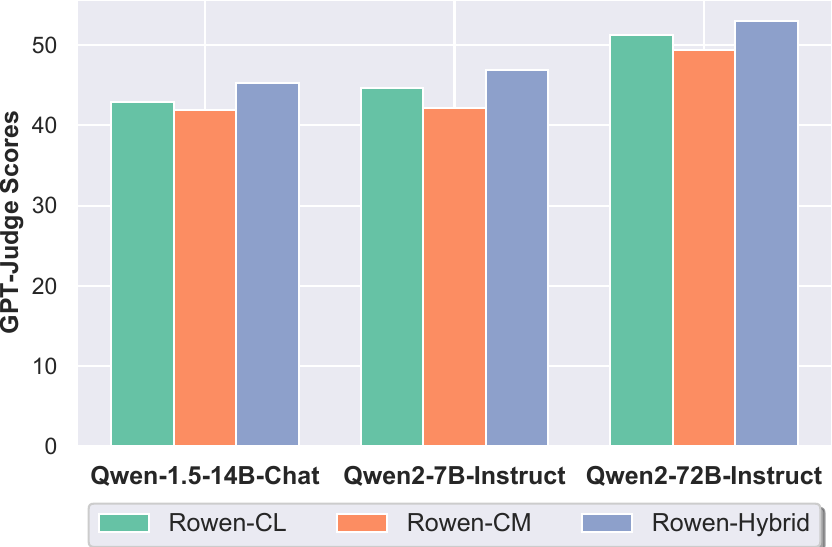}
    \caption{Experimental results of \texttt{Rowen} with open-source LLMs on TruthfulQA.}
    \label{fig:openllm}
    \vspace{-0.5cm}
\end{figure}

\paragraph{Scalability to Open-Source LLMs}
In addition to ChatGPT, we also assess Rowen's effectiveness when employing open-source language models: \texttt{Qwen1.5-14B-Chat}, \texttt{Qwen2-7B-Instruct}, and \texttt{Qwen2-72B-Instruct}\footnote{\url{https://huggingface.co/Qwen}}. These models are chosen for the cross-language detection model $\mathcal{M}$ due to their strong capabilities in following Chinese instructions, a critical feature for effective cross-language detection. The Llama-series models are not considered due to their weaker performance in generating Chinese responses. Instead, the Qwen-series models have demonstrated state-of-the-art performance in Chinese, which aligns with our research goals. For verifier LM $\mathcal{M}_{\mathcal{V}}$, we choose to use \texttt{Llama-3-8B-Instruct}\footnote{\url{https://huggingface.co/meta-llama/Meta-Llama-3-8B-Instruct}}.

Figure~\ref{fig:openllm} shows the results of three variants of \texttt{Rowen} on three open-source models on TruthfulQA dataset. Our \texttt{Rowen} methods achieve strong performance when applied to open-source LLMs, nearly matching the baseline results on ChatGPT. Specifically, the Multi-agent Debate baseline scores 50.83, while the Detect-and-Mitigate baseline scores 49.98. These findings further prove the effectiveness and scalability of our proposed \texttt{Rowen} method within open-source LLMs.

\paragraph{Scalability to Other Datasets}
We assess the Rowen model's scalability by examining its performance on datasets with answers of intermediate length, namely TriviaQA and Natural Questions, given the diverse answer lengths in TruthfulQA (long answers) and StrategyQA (binary responses). We conduct comparisons against two strong adaptive retrieval baselines, reporting the metrics of EM (Exact Match) and F1 score on their dev sets. The experimental results are shown in Table~\ref{tab:new_datasets}.

In short, our \texttt{Rowen} model, with its different variants, consistently outperforms the baselines on both datasets, suggesting a notably more robust capability in handling intermediate-length answers. Particularly, \texttt{Rowen-Hybrid} emerges as the most effective, achieving the highest scores across both metrics and datasets. This denotes a significant enhancement over baseline methods.

\begin{table}[t]
\centering
\resizebox{0.9\linewidth}{!}{
\begin{tabular}{lrrrr}
\toprule
\multirow{2}{*}{\textbf{Methods}} & \multicolumn{2}{c}{\textbf{NQ}} & \multicolumn{2}{c}{\textbf{TriviaQA}} \\
\cline{2-3}\cline{4-5}
 & EM         & F1        & EM            & F1           \\
\midrule
FLARE                    & 32.50      & 43.91     & 59.00         & 68.34        \\
Adaptive-RAG             & 35.04      & 48.44     & 58.00         & 68.97        \\
\texttt{Rowen-CL}                 & 38.08      & 55.81     & 66.50         & 76.66        \\
\texttt{Rowen-CM}                 & 37.36      & 53.27     & 65.00         & 74.98        \\
\texttt{Rowen-Hybrid}             & \textbf{39.98}      & \textbf{57.31}     & \textbf{69.04}         & \textbf{78.46}       \\
\bottomrule
\end{tabular}}
\caption{Hallucinations mitigation performance on the NQ dataset and TriviaQA dataset.}
\label{tab:new_datasets}
\vspace{-0.7cm}
\end{table}

\begin{figure*}[t]
    \centering
    \subfloat[Effect of various detection threshold.\label{fig:threshold_analysis}]{
        \includegraphics[width=0.31\textwidth]{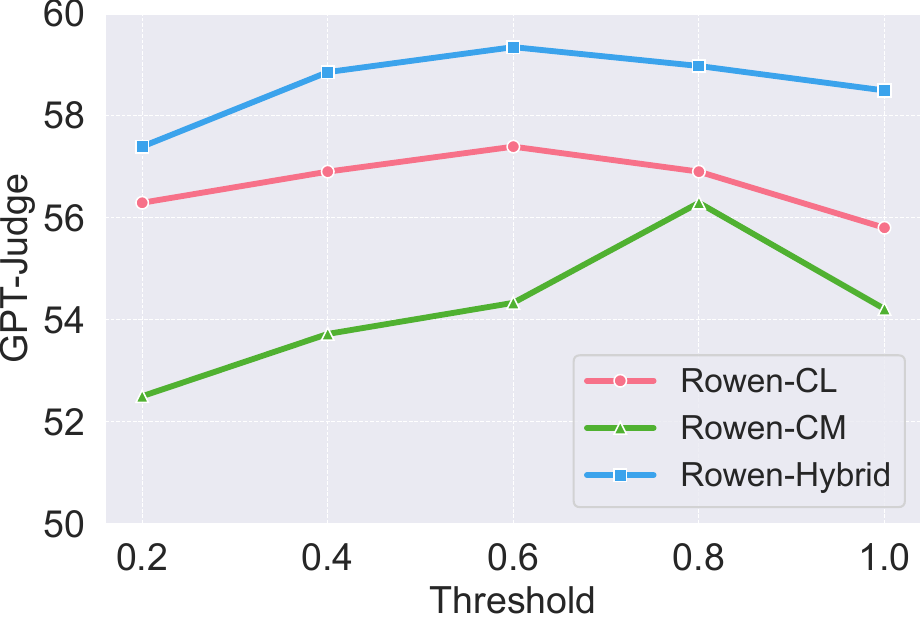}
    }
    \hspace{0.01\textwidth} 
    \subfloat[Effect of number of perturbed questions.\label{fig:perturation_analysis}]{
        \includegraphics[width=0.31\textwidth]{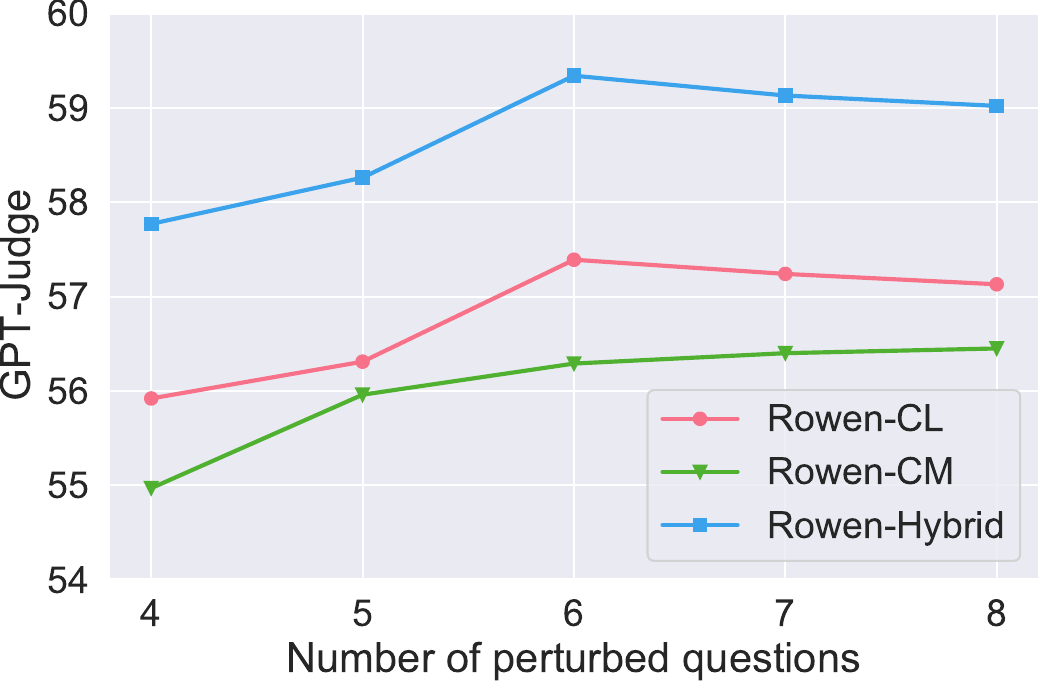}
    }
    \hspace{0.01\textwidth} 
    \subfloat[Effect of cross-model detection weight factor for \texttt{Rowen-Hybrid}\label{fig:alpha_analysis}]{
        \includegraphics[width=0.31\textwidth]{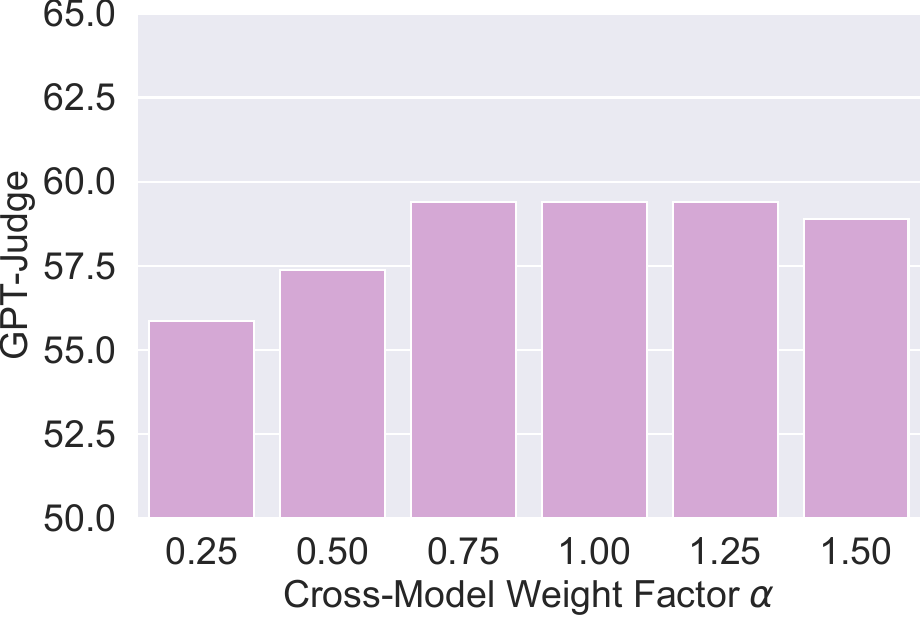}
    }
    \caption{Impact of key factors for hallucination mitigation on the TruthfulQA dataset.}
    \label{fig:combined_analysis}
    \vspace{-0.5cm}
\end{figure*}

\subsection{Impact of Hyper-parameters in \texttt{Rowen} (RQ4)}
\label{sec:hyperparameter}
We analyze the impact of key factors for hallucination mitigation on the TruthfulQA dataset, including detection thresholds, numbers of perturbed questions, and cross-model detection weights, as shown in Figure~\ref{fig:combined_analysis}.

\paragraph{Detection Threshold} We experiment with various consistency thresholds to investigate their impact on hallucination mitigation effectiveness. As shown in Figure~\ref{fig:combined_analysis}(a), we observe significant performance improvement when the detection threshold increases from 0.2 to 0.6, as this identifies more hallucinations and enhances factual accuracy by retrieving more external information. However, further increasing the threshold introduces more noise and irrelevant evidence, degrading performance. This suggests that an appropriate threshold is crucial for balancing detection accuracy and the retrieval process. Additionally, \texttt{Rowen-CL} and \texttt{Rowen-CM} have different optimal thresholds of 0.6 and 0.8, respectively, and the \texttt{Rowen-Hybrid} method demonstrates the most stable performance across varying thresholds.  

\paragraph{Number of Perturbed Questions} We also study the effect of the number of perturbed question on hallucination mitigation and the results are shown in Figure~\ref{fig:combined_analysis}(b). We observe that the performance of \texttt{Rowen} improves with an increasing number of question samples, yet the performance gain gradually diminishes after surpassing 6 question samples. This indicates that, in practice, using 6 question samples could achieve reasonably good performance at a relatively low computational cost.  

\paragraph{Cross-Model Detection Weight} We also examine the impact of different cross-model detection weight for \texttt{Rowen-Hybrid}. Figure~\ref{fig:combined_analysis}(c) shows that GPT-Judge scores significantly increase as the weight rises from 0.25 to 1.0, indicating that introducing cross-model checking can improve the accuracy of hallucination detection. However, when the weight factor exceeds 1.0, performance declines, with scores dropping to around 56 at a weight of 1.5. This suggests that while a moderate weight enhances factuality, excessively a high value allows the cross-model detection module to exert too much influence, reducing effectiveness.

\begin{figure}[t]
    \small
    \centering
    \includegraphics[width=\columnwidth]{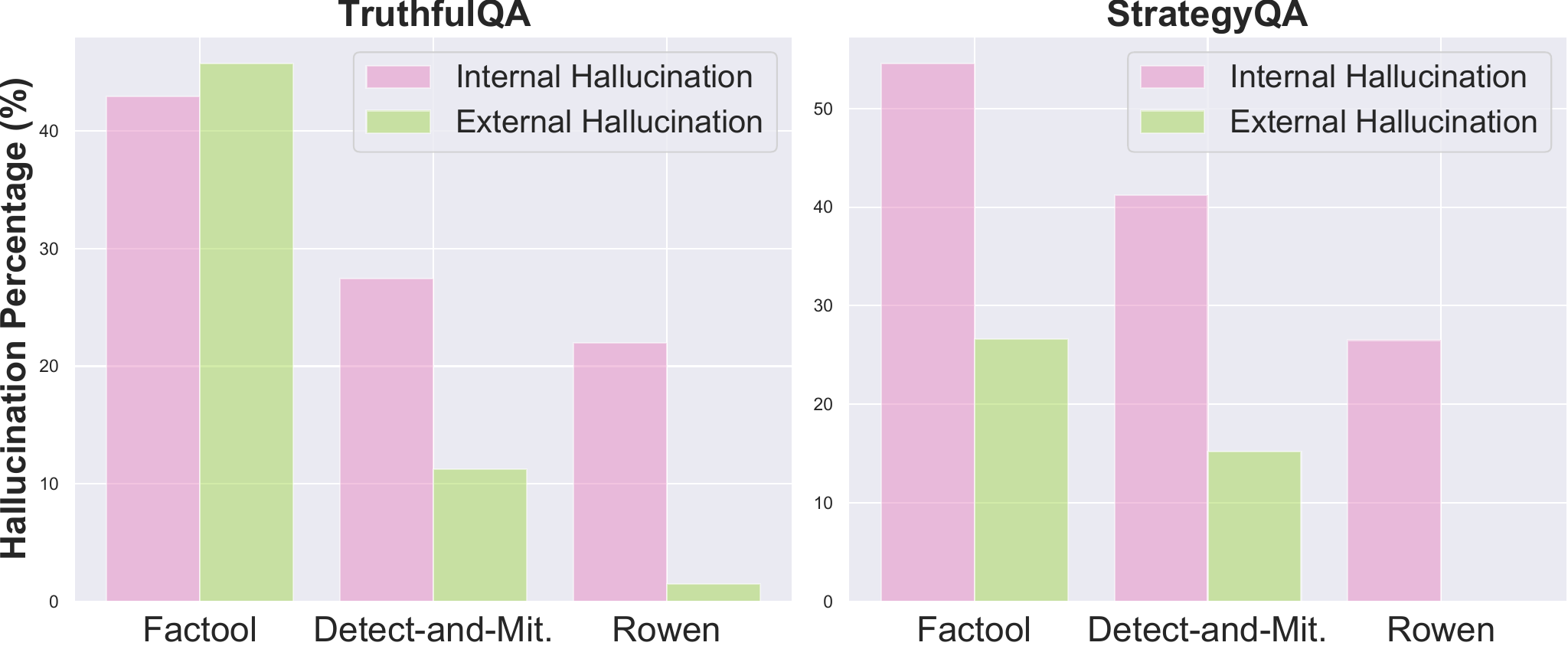}
    \caption{The percentage of internal and external hallucinations across different methods on the both dataset.}
    \label{fig:statistics}
    \vspace{-0.5cm}
\end{figure}

\paragraph{Impact of the Choice of Target Language}
\label{sec:language_pair}
To investigate how different language pairs affect the effectiveness of hallucination mitigation for \texttt{Rowen-CL}, we conduct experiments on the TruthfulQA dataset  with various target languages. The results of these experiments are presented in Table~\ref{tab:language}. It is observed that the combination of English as the source language and German as the target language yields the least favorable results. This may be attributed to their shared Germanic language family roots, which results in numerous linguistic similarities and overlapping cultural references. Conversely, as the cultural divergence between the source and target languages widens, we witness an enhancement in the performance of hallucination mitigation. This trend substantiates the hypothesis that cultural disparities between languages play a pivotal role in identifying hallucinations and bolstering the factuality of the generated responses.

\begin{table}[t]
\centering
\resizebox{0.8\linewidth}{!}{
\begin{tabular}{lcc}
\toprule
\textbf{Language Pair} & \textbf{Discrepancy} & \textbf{GPT-Judge} \\
\midrule
English - German         & Low                & 50.55             \\
English - French         & Medium              & 52.26              \\
English - Chinese        & High                 & 57.39             \\
\bottomrule
\end{tabular}}
\caption{Comparison of hallucination mitigation performance for \texttt{Rowen-CL} under different language choices on the TruthfulQA dataset.}
\label{tab:language}
\vspace{-0.5cm}
\end{table}

\begin{table}[t]
\centering
\resizebox{0.8\linewidth}{!}{
\begin{tabular}{lcc}
\toprule
\textbf{Verifier LM} & \textbf{Capability} & \textbf{GPT-Judge} \\
\midrule
Qwen-Turbo         & Low                & 52.50             \\
Qwen-Plus         & Medium              & 54.09              \\
Qwen-Max         & High                 & 56.29             \\
\bottomrule
\end{tabular}}
\caption{Comparison of hallucination mitigation performance for \texttt{Rowen-CM} under different verifier LM on the TruthfulQA dataset.}
\label{tab:model}
\vspace{-0.6cm}
\end{table}

\paragraph{Impact of the Choice of Verifier LM}
\label{sec:verifier_lm}
To investigate the impact of different verifier LMs on the efficacy of hallucination mitigation for \texttt{Rowen-CM}, we conducted experiments using various verifier LMs on the TruthfulQA dataset. The experimental results presented in Table~\ref{tab:model} demonstrate a significant improvement in hallucination mitigation as the capability of the verifier LM increases. 
Specifically, Qwen-Max exhibited the highest effectiveness in this task, achieving an efficacy of 56.29\%, while Qwen-Turbo and Qwen-Plus achieved 52.50\% and 54.09\%, respectively. 
These findings underscore the critical role of verifier LM selection in enhancing the effectiveness of hallucination mitigation tasks.

\subsection{Quantitative Analysis (RQ5)}
\label{sec:quantitative}
To verify whether \texttt{Rowen} effectively reduces internal and external hallucinations, we present a comparative analysis of the prevalence of both types of hallucinations across three methods—Factool, Detect-and-Mitigate, and \texttt{Rowen-Hybrid}.
Internal hallucinations refer to the generation of wrong responses using only the parameterized knowledge of LLMs, while external hallucinations refer to the generation of incorrect responses using noisy documents introduced after retrieval.
Figure~\ref{fig:statistics} shows that Factool and Detect-and-Mitigate are significantly prone to both types of hallucinations. Both baseline methods have high levels of internal and external hallucinations, struggling with internal coherence and external fact alignment. In contrast, \texttt{Rowen-Hybrid} effectively reduces external hallucinations by timely integrating external knowledge, thereby avoiding unnecessary information retrieval and mitigating potential errors.

\begin{table}[t]
\centering
\resizebox{0.9\linewidth}{!}{
\begin{tabular}{lrrrr}
\toprule
\multirow{2}{*}{\textbf{Methods}} & \multicolumn{2}{c}{\textbf{TruthfulQA}} & \multicolumn{2}{c}{\textbf{StrategyQA}}  \\
\cline{2-3}\cline{4-5}
& GPT-Judge & \# Calls & Accuracy & \# Calls \\
\midrule
Self-Reflection & 42.99 & 6 & 62.40 & 5 \\
Multi-agent Deb. & 50.83 & 6 & 65.73 & 6 \\
\texttt{Rowen-CL} & 57.39 & 6 & 74.00 & 5 \\
\texttt{Rowen-CM} & 56.29 & 5 & 72.40 & 4 \\
\texttt{Rowen-Hybrid} & \textbf{59.34} & 8 & \textbf{75.60} & 6 \\
\bottomrule
\end{tabular}}
\caption{Analysis on LLM calls efficiency and hallucination mitigation performance across different methods.}
\label{tab:api_calls}
\vspace{-0.5cm}
\end{table}

\begin{table}[t]
    \small
    \centering
    \resizebox{0.8\linewidth}{!}{
    \begin{tabular}{lrr}
    \toprule
    \multirow{2}{*}{\textbf{Methods}} & \multicolumn{2}{c}{\textbf{\# Num of Retrieval Calls}} \\
    \cline{2-3}
    & TruthfulQA       &  StrategyQA      \\
    \midrule
    Factool         &     12.5  & 11.6  \\
    Detect-and-Mit. &     7.2   &  5.5    \\
    FLARE           &     2.1   &   3.9   \\
    Adaptive-RAG    &     \textbf{0.9}   &   1.4   \\
    \texttt{Rowen-Hybrid}    & 1.5 & \textbf{0.5}  \\
    \bottomrule
    \end{tabular}}
    \caption{Statistics on the average number of retrieval calls to answer each question.}
    \label{tab:api_count}
    \vspace{-0.7cm}
\end{table}

\subsection{Analysis of Computation Cost (RQ6)}
\paragraph{Efficiency Analysis of LLM Calls}
To evaluate the efficiency of LLM calls in \texttt{Rowen}, we compare various methods—including Self-Reflection, Multi-agent Debate and three different \texttt{Rowen} variants—by analyzing the number of API calls required to answer a single question and their effectiveness in mitigating hallucinations. From the results shown in Table~\ref{tab:api_calls}, we find that \texttt{Rowen} variants, particularly \texttt{Rowen-CL} and \texttt{Rowen-CM}, achieve significantly better hallucination mitigation compared to the baseline methods, while using a similar number of API call.

\paragraph{Efficiency Analysis of Retrieval Calls}
To verify the retrieval efficiency of Rowen, in Table~\ref{tab:api_count}, we compare the average number of retrieval calls made by Factool,  Detect-and-Mitigate, FLARE, and \texttt{Rowen-Hybrid} to answer a question across two datasets.
Factool generates the most retrieval API calls due to verifying each claim. Detect-and-Mitigate and FLARE identify low-confident concepts in LLM outputs and call search APIs with fewer API calls. Adaptive-RAG underestimates the difficulty of adversarial questions in the TruthfulQA dataset, resulting in poor truthfulness scores despite using the minimum number of retrievals.
\texttt{Rowen-Hybrid}, which retrieves information only for uncertain responses, excels in both retrieval efficiency and factual accuracy, showcasing its superiority over other RAG methods.
This demonstrates the superiority of \texttt{Rowen-Hybrid} over other RAG methods in terms of both efficiency and effectiveness.

\section{Conclusion and Future Work}
In this paper, we introduce \texttt{Rowen}, a novel method designed to actively identify potential hallucinated responses in LLMs and correcting factual errors through an adaptive retrieval process. Our experiments on two datasets demonstrate the effectiveness of our approach. We also conduct detailed analytical experiments to underscore the impact of each module and the choices of hyper-parameters. Moving forward, we plan to expand our work by exploring how to effectively utilize retrieved evidence, even when irrelevant documents are integrated, and devise an effective method to minimize the influence of knowledge conflicts between internal knowledge and retrieved documents.

\section*{Acknowledgement}
This work was supported by the Strategic Priority Research Program of the CAS under Grants No.XDB0680302, the National Natural Science Foundation of China (NSFC) under Grants No. 62276248, the Key Research and Development Program of Xinjiang Uyghur Autonomous Region Grant No. 2024B03026, the Beijing Nova Program under Grants No. 20250484765, and the Youth Innovation Promotion Association CAS under Grants No. 2023111.

\bibliographystyle{ACM-Reference-Format}
\balance
\bibliography{sample-base}

\end{document}